\documentclass[runningheads]{llncs}

\usepackage{eccv}

\usepackage{eccvabbrv}

\usepackage{graphicx}
\usepackage{booktabs}
\usepackage{multirow}
\usepackage{adjustbox} 
\usepackage[accsupp]{axessibility}  
\usepackage{amsmath,amssymb,amsfonts}
\usepackage{mathtools}
\usepackage{lmodern}
\usepackage{microtype}
\usepackage[numbers,sort&compress]{natbib}
\usepackage{caption}
\usepackage{makecell}
\usepackage{marvosym}

\newcommand{\corrauth}{\textsuperscript{\textrm{\Letter}}}

\usepackage{hyperref}

\usepackage{orcidlink}

\usepackage[nameinlink,capitalize,noabbrev]{cleveref}

\spnewtheorem{assumption}{Assumption}{\bfseries}{\itshape}
\crefname{assumption}{Assumption}{Assumptions}
\Crefname{assumption}{Assumption}{Assumptions}

\begin{document}

\title{TecoPrompt: Temporal-Conservative Prompt Learning for Vision-Language Models} 

\titlerunning{TecoPrompt}

\author{
	Zeyi Shao\inst{1}
	\and
	Haowen Hua\inst{1}
	\and
	Jiaxin Zhang\inst{2}
	\and
	John See\inst{3}
	\\
	Zeyd Boukhers\inst{4}
	\and
	Cong Yang\inst{1}\corrauth
}

\authorrunning{Z.~Shao et al.}

\institute{
	Soochow University, Suzhou, China
	\and
	NVIDIA, Shanghai, China
	\and
	Heriot-Watt University Malaysia, Putrajaya, Malaysia
	\and
	Fraunhofer FIT, Sankt Augustin, Germany\\
	\corrauth~Corresponding author: \email{cong.yang@suda.edu.cn}
}

\maketitle

\begin{abstract}
Prompt learning adapts vision-language models, such as CLIP, by adjusting a small set of context tokens. However, under few-shot supervision, even moderate label noise can disrupt prompt optimization. To address this issue, we propose TecoPrompt, a closed-loop robust prompt-learning framework that revisits optimal transport (OT) pseudo-labeling from a temporal perspective. TecoPrompt employs an entropic OT plan in the CLIP semantic space to obtain globally consistent label candidates. It verifies the reliability of these candidates by examining trajectory stability: a noisy label is only rewritten if the OT candidate remains unchanged within a \(K\)-epoch temporal stability window and passes a confidence gate based on Exponential Moving Average (EMA). This approach helps reduce confirmation bias. The rewritten labels are then integrated back into prompt training using a tri-group objective that includes three loss functions aligned with clean, mid, and noisy subsets. Experiments on seven datasets with synthetic symmetric and asymmetric noise, as well as Food101N, demonstrate significant performance improvements. For example, on the OxfordPets dataset, with 50\% asymmetric noise, TecoPrompt achieves an accuracy of 0.843, up from 0.775.

\end{abstract}

\section{Introduction}
\label{sec:intro}
Vision-language pre-trained models (VL-PTMs) like CLIP learn to create aligned representations of images and text. This capability allows for open-vocabulary recognition and robust zero-shot transfer~\cite{radford2021clip, jia2021scaling, yu2022coca}. Building upon this alignment, prompt learning has emerged as a practical method for adaptation. It involves updating only a small set of context tokens while keeping the encoders frozen, making it efficient and data-friendly~\cite{zhou2022learning, zhou2022conditional, jia2022vpt, khattak2023maple}. Additionally, prior research~\cite{wu2023whyprompt} suggests that prompt-based adaptation may be inherently more tolerant to imperfect supervision.

Despite their successes, most current strategies~\cite{han2018coteaching,liang2022fewshot,wei2020jocor} for learning from noisy labels offer limited improvements in few-shot prompt learning, where each labeled example is highly valuable. These methods commonly utilize techniques like sample grouping, instance filtering, or robust loss functions to reduce the impact of noisy annotations. While these approaches enhance robustness, they often downweight or eliminate portions of the data, resulting in inefficient use of limited supervision. A more effective solution is label correction, which aims to convert noisy annotations into reliable targets. However, naive or overly aggressive rewriting can mistakenly alter clean labels, worsening confirmation bias, particularly when early predictions are unstable. Therefore, a high-precision and conservative correction strategy is essential to enhance supervision while carefully managing the risk of mis-correction.

\begin{figure}[t]
	\centering
	\includegraphics[width=1\linewidth]{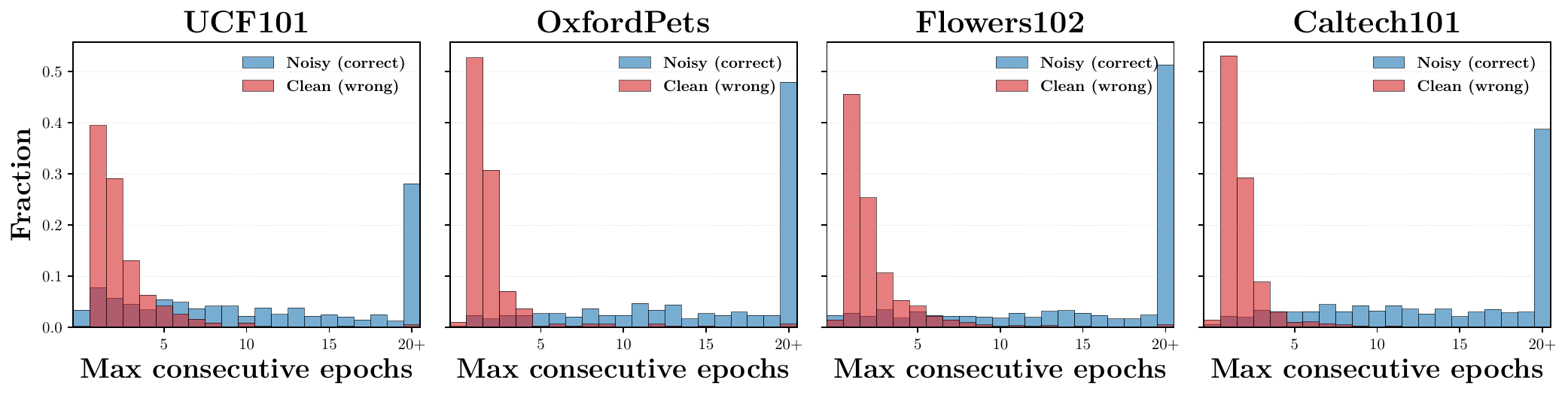}
	\caption{
		Distribution of the maximum number of consecutive epochs during which the OT pseudo-label remains unchanged.
		For noisy examples, we count runs in which the stable OT label matches the ground-truth (Noisy, correct); for clean examples, we count runs in which the stable OT label is incorrect (Clean, wrong).
		Across datasets, noisy examples frequently exhibit long stable-correct runs (including the ``$20+$'' bin), whereas clean examples rarely sustain long stable-wrong runs.}
	\label{fig:ot_temporal_consistency}
\end{figure}

Motivated by this, we revisit optimal transport (OT)-based pseudo-labeling using a globally coupled assignment. To analyze the temporal behavior of OT pseudo-labels, we conducted a controlled experiment in which 50\% random label noise was injected into the training data. In Fig.~\ref{fig:ot_temporal_consistency}, when OT pseudo-labels are recomputed across epochs with a globally coupled assignment, some samples demonstrate strong temporal consistency. Notably, noisy samples often display extended periods during which the OT label remains unchanged and aligns with the ground-truth label, while clean samples are less likely to exhibit such long, incorrect stable runs. This significant distinction suggests that OT provides reliable candidate labels, and that temporal consistency can serve as a precise criterion for determining when to implement label corrections during training.

To demonstrate when stability can be trusted for hard rectification, we analyze the supervision trajectory induced by Optimal Transport (OT) through the lens of learning dynamics~\cite{chen2023plot,xu2018maxmargin}. We consider the OT candidate distribution as a score vector and monitor the evolution of its maximum value (argmax) during training. With bounded perturbations per epoch, a sufficient margin condition ensures the invariance of the argmax. This explains why stable candidates are more reliable. Additionally, we connect OT label flips to forgetting-style dynamics. This analysis supports the use of a $K$-epoch stability window as a conservative gate and suggests that $K$ should be chosen to enhance precision in rectification~\cite{toneva2019unforgettable}.

Based on these insights, we propose TecoPrompt (Temporal-Conservative Prompt Learning), a robust and closed-loop framework for few-shot adaptation in the presence of noisy labels. TecoPrompt uses an entropically regularized optimal transport formulation to establish a globally consistent assignment between image embeddings and prompt-conditioned text prototypes. From this, we derive globally consistent candidates. To verify these candidates, TecoPrompt evaluates trajectory stability within a $K$-epoch window, alongside an exponential moving average (EMA)-based confidence gate, and performs conservative hard rectification. The corrected labels are then fed back into the prompt optimization process using a tri-group objective. This objective combines cross-entropy loss on clean and rectified samples with noise-robust losses on uncertain samples.

Succinctly, the main contributions of this work are as follows: 1) We propose TecoPrompt, a closed-loop robust prompt learning framework that couples global OT-based candidate generation with strict temporal verification for reliable label rectification. 2) We establish sufficient conditions that justify hard label rewriting and show how the stability window and confidence gating improve correction precision. 3) We conduct extensive experiments on diverse datasets under various noise settings, consistently achieving substantial performance gains over prior approaches (e.g., 66.6\% vs. 55.3\% on Flowers102 under 75\% asymmetric noise).

\section{Related Work}
\label{sec:related}

\subsubsection*{Prompt Learning for Vision-Language Models.}
Prompt learning adapts CLIP-style vision-language models by tuning lightweight prompts while keeping the backbone frozen. A representative line learns continuous textual contexts, exemplified by CoOp and the instance-conditional CoCoOp ~\cite{zhou2022learning,zhou2022conditional}. Another line extends prompting beyond pure text to strengthen cross-modal adaptation, such as visual prompt tokens and joint prompting across modalities ~\cite{jia2022vpt,khattak2023maple}. Recent studies further improve prompt transfer under distribution shifts by introducing source-aware regularization ~\cite{khattak2023promptsrc} or by explicitly modeling domain adaptation for CLIP ~\cite{lu2022proda}. In parallel, label-free test-time prompt tuning updates prompts using unlabeled target data at inference, including TPT and its calibrated variant, C-TPT ~\cite{shu2022tpt,yoon2024ctpt}. Beyond these core directions, prompt learning is increasingly studied under alternative training regimes that reduce annotation dependence or enhance generalization, such as unsupervised prompt distillation ~\cite{li2024promptkd} and global-local prompting ~\cite{lafon2024gallop}, which leverage both holistic and region-level cues.

\subsubsection*{Learning with Noisy Labels.}
Noisy supervision often induces memorization and biased optimization in deep networks ~\cite{arpit2017memorization,zhang2017rethinking,chen2023twowrongs}.
Common defenses include robust training dynamics and regularization ~\cite{xia2021robustearly,lyu2020curriculum,hendrycks2019pretrain},
noise-robust loss design ~\cite{zhang2018generalized,feng2020can},
explicit objective correction via noise modeling or transition estimation ~\cite{patrini2017forward,yao2020dualt,zhang2021identifiable},
and data-centric selection or refinement ~\cite{li2020dividemix,reed2015bootstrapping,song2019selfie,lee2018cleannet}.

Prompt tuning's empirically observed resilience to noisy supervision was highlighted by Wu et al.~\cite{wu2023whyprompt}; JoAPR~\cite{guo2024joapr} separates clean/noisy samples via a Gaussian-mixture partitioning and label-refinement pipeline and then retrains on the purified set; NLPrompt enhances prompt learning by using PromptMAE and PromptOT. The PromptOT method purifies noisy labels by dividing the data into clean and noisy subsets. CE is applied to the clean subset, while MAE is used for noisy samples to improve robustness. In contrast, our method differs from NLPrompt in that it integrates temporal consistency into label corrections. TecoPrompt checks label stability over $K$ epochs before committing corrections, ensuring more reliable updates. Additionally, TecoPrompt employs a tri-group loss to enable more nuanced optimization.

\subsubsection*{Optimal Transport for Global Consistency.}
Optimal transport (OT) provides a principled global coupling mechanism and can be solved efficiently via entropic regularization and Sinkhorn iterations ~\cite{cuturi2013sinkhorn,peyre2019ot}.
OT has been used as a global signal for learning under noisy supervision, such as filtering/reweighting or curriculum-aware transport ~\cite{feng2023otfilter,chang2023csot}, and has also been introduced into prompt learning to exploit the shared embedding space ~\cite{chen2023plot}.
Many existing OT-based prompt-learning approaches primarily employ OT for purification, subset assignment, or loss re-weighting.
In contrast, our method treats OT assignments as structured, globally consistent label candidates, and further introduces a sample-wise temporal consistency criterion (based on the stability of historical OT assignments) to decide whether to trigger hard label rewriting, aligning with the broader principle of temporal consistency in semi-supervised learning while explicitly controlling confirmation bias ~\cite{laine2017temporal,tarvainen2017mean,chen2023twowrongs}.

\section{Formulation of Temporal-Consistency OT Label Rewriting}
In this section, we provide a theoretical characterization of OT label rewriting triggered by temporal consistency in TecoPrompt. Building on temporal consistency as a reliability prior widely used in semi-supervised learning \cite{laine2017temporal} and teacher–student consistency targets \cite{tarvainen2017mean}, we formalize when a temporally stable OT assignment can serve as a reliable label candidate and show that strengthening the temporal consistency requirement yields a more conservative yet reliable rewriting set.

\subsection{Notation.}
\label{subsec:notation}
Scalars are denoted by non-bold letters, vectors by lowercase bold letters, and matrices by uppercase bold letters. The indicator function is written as $\mathbf{1}[\cdot]$, and we use $[n]=\{1,2,\dots,n\}$. For each sample $i\in[N]$, let $y_i^\star\in[C]$ denote its (latent) clean label and $\tilde{y}_i\in[C]$ the observed (possibly noisy) label. At epoch $t$, the OT-induced pseudo-label is $\hat y_i^{(t)}\in[C]$, with a confidence score $c_i^{(t)}\in[0,1]$.

Given a temporal stability window of size $K$, we define three events to characterize temporal stability and reliability. The temporal consistency event is $\mathcal{E}_K(i)=\{\hat y_i^{(t-K+1)}=\cdots=\hat y_i^{(t)}\neq \tilde y_i\}$, meaning that the pseudo-label remains unchanged within the $K$-epoch stability window and disagrees with the observed label. The reliability event is $\mathcal{D}_K(i)=\{c_i^{(t-K+1)}\ge \theta,\ldots,c_i^{(t)}\ge \theta\}$, requiring the confidence to stay above a threshold $\theta$ throughout the same window. The rewrite-candidate event is defined as $\mathcal{A}_K(i)=\mathcal{E}_K(i)\cap \mathcal{D}_K(i)$.

\subsection{OT-Forgettability and a $2\times2$ Decomposition.}
We characterize sample-wise stability through the OT pseudo-label trajectory.
Let $\hat y_i^{(s)}\in[C]$ denote the OT-induced pseudo-label of sample $i$ at epoch $s$.
After warm-up, we monitor the trajectory over a window of length $T_w$ starting from epoch $t_0$.
We quantify temporal instability using the one-step flip indicator and its flip rate:
\begin{equation}
d_i^{(s)}=\mathbf{1}\!\left[\hat y_i^{(s+1)}\neq \hat y_i^{(s)}\right],\quad
\phi_i=\frac{1}{T_w}\sum_{s=t_0}^{t_0+T_w-1} d_i^{(s)}\in[0,1].
\end{equation}

A smaller $\phi_i$ indicates a more stable OT assignment, while a larger value indicates frequent changes.
Given a threshold $\phi_0\in(0,1)$, we define the OT-forgettability indicator as follows:
\begin{equation}
F_i=\mathbf{1}[\phi_i>\phi_0]
.
\end{equation}
We call $F_i=0$ OT-unforgettable and $F_i=1$ OT-forgettable.
Unlike the classic example-forgetting definition based on correctness with respect to $y_i^\star$,
$\phi_i$ and $F_i$ are directly observable from the OT pseudo-label trajectory.

For theoretical analysis, let $\tilde y_i$ be the observed label and $y_i^\star$ be the ground-truth label.
We introduce the latent noise indicator:
\begin{equation}
Z_i=\mathbf{1}[\tilde y_i\neq y_i^\star].
\end{equation}
which is used only to describe the clean/noisy decomposition and is not required by the rewriting rule.
The pair $(Z_i,F_i)$ induces a $2\times2$ partition:

\begin{equation}
	\begin{array}{c|c@{\hspace{12pt}}c}
		& F_i=0,\ \text{OT-unforgettable} & F_i=1,\ \text{OT-forgettable}\\
		\hline
		Z_i=0,\ \text{clean} & \mathcal{S}_{c,u} & \mathcal{S}_{c,f}\\
		Z_i=1,\ \text{noisy} & \mathcal{S}_{n,u} & \mathcal{S}_{n,f}
	\end{array}
	\label{eq:2x2}
\end{equation}

Intuitively, $\mathcal{S}_{n,u}$ contains noisy yet OT-unforgettable samples: despite incorrect observed labels, their OT pseudo-labels remain consistent across training, making them the primary targets for conservative label rewriting.

\subsubsection*{Rewriting objective.}
Our goal is to correct supervision mainly on the noisy OT-unforgettable subset $\mathcal{S}_{n,u}$.
Meanwhile, the procedure should remain conservative on clean data by requiring stable disagreement and high confidence.

\subsection{Temporal-Consistency Rewriting.}
Following Sec.~\ref{subsec:notation}, $\mathcal{A}_K(i)$ collects samples whose OT-induced pseudo-labels are temporally stable and simultaneously disagree with the observed labels.

Therefore, a sample is rewritten only if it is temporally consistent, disagrees with $\tilde y_i$, and maintains high confidence over the history window.

\begin{theorem}
\label{thm:precision}
Under the proposed assumptions, increasing the consistency window size $K$
improves the reliability of the rewritten labels.
In particular, for sufficiently large $K$, the probability of correct
label rewriting can be made arbitrarily high.
\end{theorem}

\noindent
The proof is provided in the Appendix.
This result shows that longer OT-consistency trajectories progressively
eliminate unstable samples, leading to robust label correction under noise.

\section{Method}

\begin{figure}[t]
	\centering
	\includegraphics[width=1\linewidth]{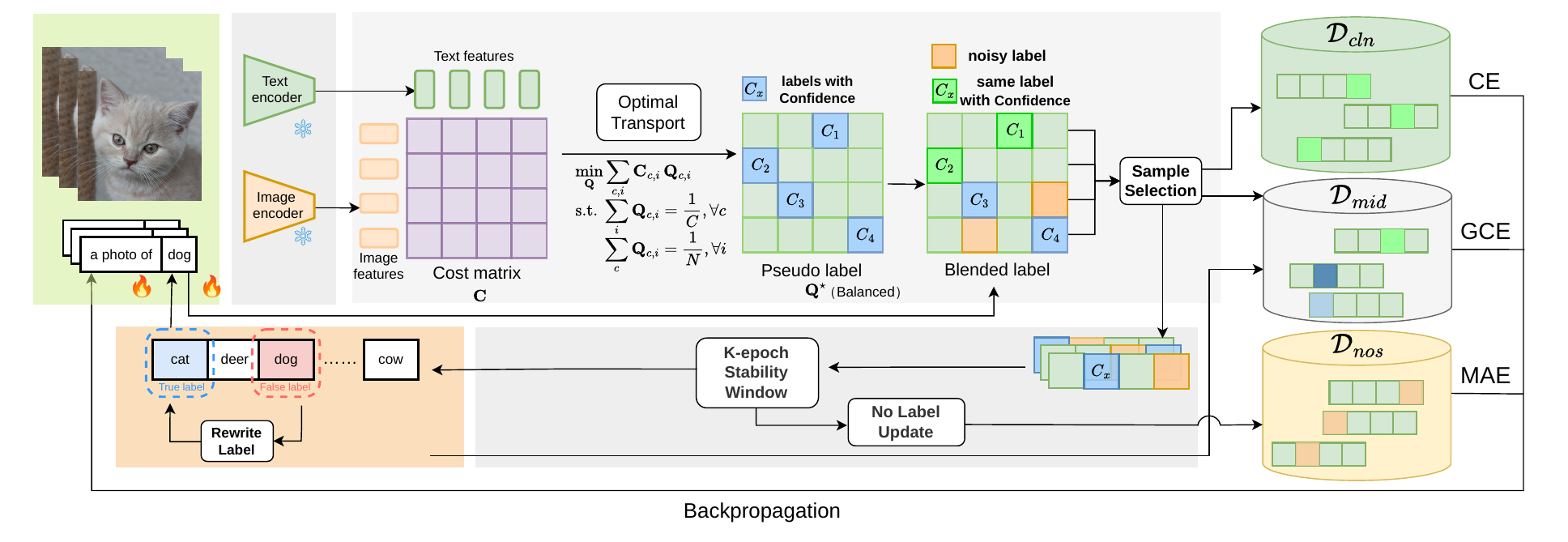}
	\caption{Overview of the proposed TecoPrompt framework. Given frozen CLIP image/text encoders, we compute image embeddings and prompt-conditioned text prototypes, and solve a balanced entropic optimal transport problem to obtain globally consistent pseudo-label distributions. We then blend the OT pseudo-labels with the observed noisy labels using confidence weighting (where confidence scores are EMA-smoothed), perform sample selection to form clean/mid/noisy groups, and optimize prompts with a tri-group objective. In parallel, we track the OT-induced label trajectory over epochs and apply EMA-gated temporal consistency with a $K$-epoch stability window to conservatively trigger hard-label rewriting, thereby forming a closed-loop refinement that mitigates confirmation bias.}
	\label{fig:figure2}
\end{figure}

As illustrated in Fig.~\ref{fig:figure2}, our core idea is to combine a global view and a sample-wise view: OT provides globally consistent label candidates, while temporal tracking identifies which candidates are sufficiently reliable to trigger conservative label rewriting. Our method consists of five components: (i) Text \& Image Encoder, (ii) OT-based purification, (iii) EMA-smoothed OT confidence, (iv) temporal consistency label rewriting, and (v) ternary grouping with loss-specific training.

\subsection{Text \& Image Encoder.} We use a CLIP-style prompt-learning setup with a frozen text encoder $h(\cdot)$ and a frozen image encoder $g(\cdot)$. For each class $c\in[C]$, we feed a learnable context prompt $p$ together with a fixed class prompt $p_c$ into the text encoder \(h\) to obtain a class-specific text feature:
\begin{equation}
	\mathbf{h}_c = h(p, p_c)\in\mathbb{R}^d. 
\end{equation}

For each input image $x_i$, we extract a $d$-dimensional image representation $\mathbf{g}_i\in\mathbb{R}^d$ using the frozen image encoder $g(\cdot)$.

We define the similarity between an image feature and a class text feature as their inner product:
\begin{equation}
	\mathrm{sim}(\mathbf{g}_i,\mathbf{h}_c)=\langle \mathbf{g}_i,\mathbf{h}_c\rangle.
\end{equation}
and obtain the predictive distribution by applying a softmax operator to the resulting similarity vector over all classes:
\begin{equation}
	\mathbf{s}_i(p)=\mathrm{softmax}\big(\mathrm{sim}(\mathbf{g}_i,\{\mathbf{h}_c\}_{c=1}^C)\big)\in\mathbb{R}^C.
\end{equation}

In implementation, we stack the class text features $\{\mathbf{h}_c\}_{c=1}^C$ and image features $\{\mathbf{g}_i\}_{i=1}^N$ into matrices $\mathbf{T}=[\mathbf{h}_1;\ldots;\mathbf{h}_C]\in\mathbb{R}^{C\times d}$ and $\mathbf{I}=[\mathbf{g}_1;\ldots;\mathbf{g}_N]\in\mathbb{R}^{N\times d}$, and compute the pairwise similarity matrix $\mathbf{S}=\mathbf{T}\mathbf{I}^{\top}\in\mathbb{R}^{C\times N}$.

\subsection{OT-based Purification}
We apply entropically regularized OT in the shared vision-language embedding space, where class-wise text features serve as prototypes and image features as samples. Solved efficiently by Sinkhorn iterations~\cite{cuturi2013sinkhorn,peyre2019ot}, OT yields balanced soft pseudo-labels for each sample, providing a lightweight purification signal for data partitioning, confidence estimation, and subsequent refinement.

Given the similarity matrix $\mathbf{S}\in\mathbb{R}^{C\times N}$ computed above, 
we construct the OT cost matrix $\mathbf{C}$ via an element-wise negative log transform of $\mathbf{S}$, 
and solve the following balanced OT problem:
\begin{equation}
	\min_{\mathbf{Q}\ge 0}\ \langle \mathbf{C}, \mathbf{Q}\rangle-\epsilon H(\mathbf{Q})
	\ \ \text{s.t.}\ \ 
	\mathbf{Q}\mathbf{1}_{N}=\tfrac{1}{C}\mathbf{1}_{C},\ 
	\mathbf{Q}^{\top}\mathbf{1}_{C}=\tfrac{1}{N}\mathbf{1}_{N}.
\end{equation}
where $H(\mathbf{Q})=-\sum_{c=1}^{C}\sum_{i=1}^{N}Q_{c,i}\log Q_{c,i}$ 
is the entropy of the transport plan, and $\epsilon>0$ controls the strength of the entropic regularization.
The optimal coupling $\mathbf{Q}^{\star}$ induces a soft pseudo-label distribution for each sample (column-wise). We denote the OT-induced pseudo-label by $\hat y_i^{(t)} \in [C]$ with confidence $c_i^{(t)}\in[0,1]$, and take:
\begin{equation}
	\hat{y}_i^{(t)}=\arg\max_{c\in[C]} Q^{\star}_{c,i}, \quad c_i^{(t)}=\max_{c\in[C]} Q^{\star}_{c,i}.
\end{equation}
as the OT signal used to guide subsequent sample selection and to quantify temporal consistency across iterations.

\subsection{EMA-Smoothed OT Confidence}
Single-epoch OT confidence can be noisy. We therefore maintain an exponential moving average (EMA) confidence for each sample:
\begin{equation}
	\bar c_i^{(t)} =
	\begin{cases}
		c_i^{(t)}, & t=1,\\
		\beta \bar c_i^{(t-1)} + (1-\beta)c_i^{(t)}, & t>1.
	\end{cases}
	\label{eq:ema_conf}
\end{equation}
where $\beta \in [0,1)$ controls the smoothing strength.
We also keep a short history of OT pseudo-labels and confidences to support temporal consistency checking.

\begin{figure}[t]
    \centering
    \includegraphics[width=1\linewidth]{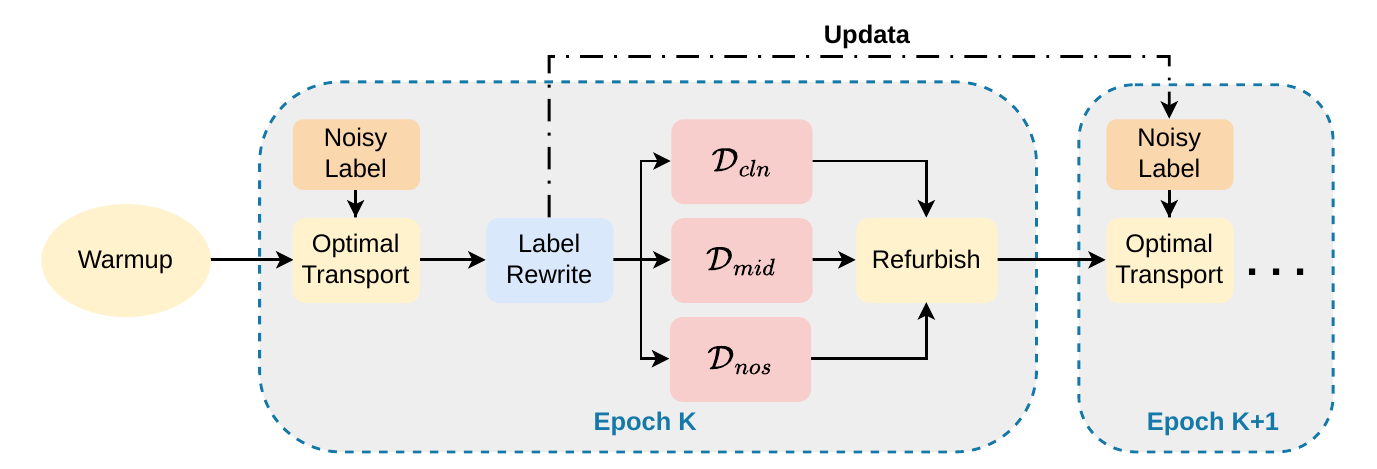}
    \caption{Optimal Transport generates pseudo-label candidates to rewrite noisy annotations. Through iterative updates over epochs, incorrect labels are progressively corrected, reducing the noise rate and improving the quality of supervision.}
    \label{fig:placeholder}
\end{figure}

\subsection{Temporal Consistency Label Rewriting}
We trigger hard label rewriting only when the OT pseudo-label is both temporally consistent and sufficiently confident, as illustrated in Fig.~\ref{fig:placeholder}.
Following Sec.~\ref{subsec:notation}, where $\mathcal{A}_K(i)$ denotes the rewrite-candidate event requiring stability within the $K$-epoch window and sustained confidence.
For eligible samples, we rewrite the observed label by the stable OT candidate:
\begin{equation}
	\tilde y_i \leftarrow \hat y_i^{(t)} \quad \text{if } \mathcal{A}_K(i) \text{ holds.}
	\label{eq:rewrite_rule}
\end{equation}
This criterion enforces a conservative rewrite policy by requiring both temporal stability and sustained confidence before updating labels.

\subsubsection*{warmup.} To avoid aggressive early corrections and reduce confirmation bias, we introduce a warmup condition. Let \(t\) denote the current epoch index and treat \(t<K\) as the warmup stage (warmup length \(=K-1\) epochs), during which we only collect the OT-induced label assignments into the temporal buffer without applying any hard rewrite. Hard OT label rewriting is enabled only once \(t\ge K\), i.e., after accumulating a full $K$-epoch stability window. In addition to preventing unstable early predictions from triggering irreversible corrections, this strategy allows the model to reach a more reliable regime and enforces multi-epoch consistency as a prerequisite for rewriting.

\subsection{Ternary Grouping and Training Objectives}
After smoothing and consistency checking, and applying label rewriting when triggered, we partition the training set $\mathcal{D}$ into three pairwise disjoint subsets $\mathcal{D}_{\text{cln}}$, $\mathcal{D}_{\text{mid}}$, and
$\mathcal{D}_{\text{nos}}$, such that
$\mathcal{D}=\mathcal{D}_{\text{cln}}\cup\mathcal{D}_{\text{mid}}\cup\mathcal{D}_{\text{nos}}$
and $\mathcal{D}_{a}\cap\mathcal{D}_{b}=\emptyset$ for $a\neq b$.

\subsubsection*{Clean set.}
A sample is placed into $\mathcal{D}_{\text{cln}}$ if its OT pseudo-label agrees with the observed label and its EMA confidence passes a strict threshold. 
Here $\tau_{\text{clean}}$ and $\tau_{\text{mid}}$ denote confidence thresholds with $\tau_{\text{clean}}>\tau_{\text{mid}}$:
\begin{equation}
	i\in \mathcal{D}_{\text{cln}}
	\quad \Longleftrightarrow \quad
	\hat y_i^{(t)}=\tilde y_i \ \ \text{and}\ \ \bar c_i^{(t)} \ge \tau_{\text{clean}}.
	\label{eq:clean_set}
\end{equation}

\subsubsection*{Mid set.}
The mid set collects samples that are likely usable but not fully reliable. Concretely, we include either agreement samples with moderate EMA confidence or  rewrite-eligible samples not already assigned to $\mathcal{D}_{\text{cln}}$:
\begin{equation}
	i\in \mathcal{D}_{\text{mid}}
	\quad \Longleftrightarrow \quad
	\Big(\hat y_i^{(t)}=\tilde y_i \ \ \text{and}\ \ \tau_{\text{mid}}\le \bar c_i^{(t)} < \tau_{\text{clean}}\Big)
	\ \ \text{or}\ \
	\mathcal{A}_K(i).
	\label{eq:mid_set}
\end{equation}

\subsubsection*{Noisy set.}
All remaining samples are treated as noisy:
\begin{equation}
	\mathcal{D}_{\text{nos}} = \mathcal{D}\setminus\left(\mathcal{D}_{\text{cln}}\cup \mathcal{D}_{\text{mid}}\right).
	\label{eq:noisy_set}
\end{equation}

\subsubsection*{Loss functions.}

Let $\mathbf{s}_i\in\Delta^{C-1}$ denote the predicted class distribution obtained from the similarity scores via a softmax operator, and let $s_{i,\tilde y_i}$ denote the probability assigned to the current training label $\tilde y_i$.
We optimize a weighted sum of losses over the three subsets:
\begin{equation}
	\mathcal{L}
	= \lambda_{\mathrm{c}}\mathcal{L}_{\text{cln}}
	+ \lambda_{\mathrm{m}}\mathcal{L}_{\text{mid}}
	+ \lambda_{\mathrm{n}}\mathcal{L}_{\text{nos}}.
	\label{eq:total_loss}
\end{equation}
where $\lambda_{\mathrm{c}},\lambda_{\mathrm{m}},\lambda_{\mathrm{n}}\ge 0$ control the relative contribution of each term.
Our design follows a ``strong-to-robust'' principle based on subset reliability:
The clean subset is expected to be highly reliable and thus benefits from the strong discriminability of cross-entropy (CE);
The mid subset may contain mild corruption, so we adopt generalized cross-entropy (GCE) with $q\in(0,1)$ as a compromise that down-weights low-confidence samples while preserving learning efficacy;
The noisy subset is potentially heavily corrupted, for which we use an MAE-style objective that is less sensitive to mislabeled targets and helps prevent overfitting to noise:

\begin{equation}
\mathcal{L}_{\text{cln}} = \sum_{i\in \mathcal{D}_{\text{cln}}}\!\! -\log s_{i,\tilde y_i}, \ 
\mathcal{L}_{\text{mid}} = \sum_{i\in \mathcal{D}_{\text{mid}}}\!\! \frac{1-s_{i,\tilde y_i}^{\,q}}{q}, \ 
\mathcal{L}_{\text{nos}} = \sum_{i\in \mathcal{D}_{\text{nos}}}\!\! (1-s_{i,\tilde y_i}).
\label{eq:loss_terms}
\end{equation}

\section{Experiments}
\label{sec:experiments}

We conduct extensive experiments to evaluate the effectiveness and robustness of TecoPrompt under few-shot prompt learning with noisy supervision.

\subsection{Benchmarks and Baselines}
\label{subsec:datasets_baselines}

\noindent\textbf{Synthetic noisy benchmarks.}
We use seven widely adopted visual classification datasets: Flowers102~\cite{nilsback2008flowers}, DTD~\cite{cimpoi2014dtd}, EuroSAT~\cite{helber2019eurosat}, OxfordPets~\cite{parkhi2012oxfordpets}, StanfordCars~\cite{krause2013stanfordcars}, UCF101~\cite{soomro2012ucf101}, and Caltech101~\cite{fei2004caltech101}. These datasets are originally clean; we corrupt labels only on the training split (Sec.~\ref{subsec:impl_details}) and keep the official test split unchanged for evaluation, as in~\cite{pan2025nlprompt}. For each dataset, we construct a 16-shot training set per class.

\noindent\textbf{Real-world noisy benchmark.}
We additionally evaluate on Food101N~\cite{lee2018cleannet}, a real-world dataset with naturally noisy web-collected annotations, to assess robustness under practical noise patterns beyond controlled synthetic label flips.

\noindent\textbf{Baselines.}
We compare TecoPrompt with CoOp~\cite{zhou2022learning}, CoOp trained with generalized cross-entropy (CoOp+GCE)~\cite{wu2023whyprompt}, JoAPR~\cite{guo2024joapr}, and NLPrompt. All methods use the same backbone and few-shot splits for fair comparison.

\subsection{Implementation Details}
\label{subsec:impl_details}

We evaluate two corruption patterns following~\cite{pan2025nlprompt} and apply noise only to the training labels.
Under symmetric noise (Sym), each label is independently flipped to a uniformly sampled incorrect class with probability $\eta$.
Under asymmetric noise (Asym), we use a fixed class-dependent mapping, in which each class is flipped to a designated successor class with probability $\eta$, thereby producing structured corruption where noisy labels are often less distinguishable, and the setting is generally more challenging.

We adopt the same experimental setup as NLPrompt~\cite{pan2025nlprompt} for fair comparison. All experiments are conducted with the pre-trained CLIP model~\cite{radford2021clip} using ResNet-50 as the image encoder. The text encoder is the CLIP text transformer, and we freeze both encoders while optimizing only the learnable prompt parameters.

We train models for 200 epochs using SGD with an initial learning rate of 0.002 and a cosine annealing schedule. For prompt design, we use 16 shared context tokens and place the class token at the end of the prompt, following the standard CoOp-style template~\cite{zhou2022learning}. All experiments were conducted on a cluster equipped with NVIDIA A100 GPUs using PyTorch~\cite{paszke2019pytorch}. All results are reported as the average accuracy over three random seeds, and the highest accuracy in each column is highlighted in bold.
\begin{table*}[t]
\caption{Classification performance (accuracy, \%) under symmetric (Sym) and asymmetric (Asym) label noise. Except for TecoPrompt, all results are taken from NLPrompt~\cite{pan2025nlprompt}.}
\label{tab:noise_results_with_Tecoprompt}

\centering
\fontsize{14pt}{14.5pt}\selectfont
\setlength{\tabcolsep}{2pt}
\renewcommand{\arraystretch}{1.15}

\begin{adjustbox}{width=\textwidth,center}
\begin{tabular}{c|c|cccccc|cccccc}
\Xhline{1.5pt}
\multirow{2}{*}{Dataset} & \multirow{2}{*}{Method}
& \multicolumn{6}{c|}{Noise Rate: Sym}
& \multicolumn{6}{c}{Noise Rate: Asym} \\
& & 12.5\% & 25.0\% & 37.5\% & 50.0\% & 62.5\% & 75.0\%
& 12.5\% & 25.0\% & 37.5\% & 50.0\% & 62.5\% & 75.0\% \\
\Xhline{1pt}

\multirow{5}{*}{Flowers102}
& CoOp     & 88.9 & 83.5 & 77.9 & 70.1 & 55.6 & 37.2 & 87.0 & 74.7 & 60.4 & 42.6 & 26.5 & 12.6 \\
& GCE      & 88.8 & 88.3 & 86.7 & 84.1 & 78.4 & 70.4 & 88.4 & 86.4 & 80.3 & 69.9 & 61.5 & 39.2 \\
& JoAPR    & 85.6 & 81.2 & 74.6 & 70.2 & 67.9 & 66.9 & 85.2 & 79.6 & 74.0 & 73.8 & 53.4 & 13.3 \\
& NLPrompt & 93.9 & 92.6 & 92.7 & 89.9 & 84.8 & 76.8 & 93.8 & 93.4 & 91.8 & 81.1 & 73.6 & 55.3 \\
& TecoPrompt
& \textbf{94.0} & \textbf{93.7} & \textbf{92.9} & \textbf{90.5} & \textbf{86.1} & \textbf{82.3}
& \textbf{94.1} & \textbf{93.5} & \textbf{92.2} & \textbf{89.4} & \textbf{80.4} & \textbf{66.6} \\
\Xhline{1pt}

\multirow{5}{*}{DTD}
& CoOp     & 56.0 & 49.6 & 43.3 & 34.4 & 27.8 & 17.3 & 55.6 & 47.8 & 38.1 & 29.6 & 20.5 & 11.7 \\
& GCE      & 61.0 & 59.8 & 56.8 & 50.7 & 43.6 & 33.7 & 60.7 & 57.6 & 52.7 & 44.0 & 33.4 & 18.2 \\
& JoAPR    & 58.1 & 57.7 & 56.3 & 53.0 & 48.1 & 29.9 & 52.4 & 56.6 & 53.1 & 48.9 & 40.2 & 28.3 \\
& NLPrompt & \textbf{63.0} & 61.2 & 59.2 & 55.2 & 49.0 & 39.8 & 62.3 & 60.6 & 56.5 & 50.8 & 40.3 & 28.4 \\
& TecoPrompt
& 62.7 & \textbf{61.8} & \textbf{59.8} & \textbf{58.1} & \textbf{54.2} & \textbf{49.8}
& \textbf{62.8} & \textbf{62.7} & \textbf{58.3} & \textbf{56.5} & \textbf{48.5} & \textbf{38.4} \\
\Xhline{1pt}

\multirow{5}{*}{EuroSAT}
& CoOp     & 76.5 & 69.2 & 61.7 & 52.3 & 37.6 & 26.7 & 76.0 & 66.3 & 53.8 & 41.2 & 28.0 & 17.4 \\
& GCE      & 82.1 & 78.6 & 74.7 & 63.1 & 49.7 & 31.4 & 78.2 & 72.7 & 63.6 & 45.3 & 22.9 & 12.1 \\
& JoAPR    & 75.1 & 61.1 & 60.9 & 63.6 & 39.0 & 27.3 & 69.4 & 67.3 & 59.4 & 47.6 & 33.9 & 17.5 \\
& NLPrompt & 82.5 & 79.5 & 78.1 & 66.7 & 63.5 & 43.8 & 80.1 & 77.1 & 71.4 & 54.3 & 37.3 & 32.7 \\
& TecoPrompt
& \textbf{83.5} & \textbf{80.7} & \textbf{78.9} & \textbf{72.7} & \textbf{68.7} & \textbf{52.1}
& \textbf{81.1} & \textbf{80.1} & \textbf{78.7} & \textbf{63.7} & \textbf{37.9} & \textbf{33.7} \\
\Xhline{1pt}

\multirow{5}{*}{OxfordPets}
& CoOp     & 76.5 & 66.7 & 60.3 & 47.0 & 35.8 & 24.6 & 76.1 & 66.2 & 52.5 & 38.7 & 26.6 & 14.9 \\
& GCE      & 85.6 & 84.6 & 83.7 & 79.2 & 71.4 & 53.2 & 85.5 & 83.0 & 76.7 & 68.1 & 50.7 & 32.0 \\
& JoAPR    & 84.0 & 83.3 & 83.2 & 83.1 & 82.4 & 74.4 & 82.9 & 83.4 & 79.1 & 75.8 & 52.7 & 43.6 \\
& NLPrompt & 86.2 & 86.0 & 85.3 & 84.9 & 83.6 & 70.8 & 86.0 & 85.0 & 82.4 & 77.5 & 66.3 & 48.6 \\
& TecoPrompt
& \textbf{86.7} & \textbf{86.3} & \textbf{85.7} & \textbf{85.1} & \textbf{84.2} & \textbf{84.6}
& \textbf{86.7} & \textbf{85.2} & \textbf{84.9} & \textbf{84.3} & \textbf{83.0} & \textbf{75.3} \\
\Xhline{1pt}

\multirow{5}{*}{StanfordCars}
& CoOp     & 66.2 & 59.7 & 53.4 & 45.9 & 35.7 & 22.9 & 65.8 & 57.1 & 46.2 & 33.7 & 22.4 & 12.8 \\
& GCE      & \textbf{69.7} & 66.4 & 66.5 & 63.8 & 59.3 & 50.9 & 70.0 & 66.5 & 61.2 & 53.7 & 39.7 & 26.6 \\
& JoAPR    & 68.6 & 66.3 & 62.8 & 56.7 & 48.5 & 39.4 & 66.5 & 61.7 & 51.5 & 42.0 & 30.8 & 23.0 \\
& NLPrompt & 69.4 & 68.8 & 67.2 & 65.6 & 62.8 & 58.3 & 69.8 & 67.5 & 64.2 & 59.0 & 50.9 & 39.5 \\
& TecoPrompt
& \textbf{69.7} & \textbf{68.9} & \textbf{68.2} & \textbf{66.8} & \textbf{64.4} & \textbf{60.4}
& \textbf{70.2} & \textbf{68.7} & \textbf{66.7} & \textbf{63.2} & \textbf{58.2} & \textbf{51.9} \\
\Xhline{1pt}

\multirow{5}{*}{UCF101}
& CoOp     & 69.0 & 63.4 & 58.2 & 49.7 & 40.8 & 26.3 & 67.2 & 58.1 & 46.5 & 34.4 & 23.7 & 13.2 \\
& GCE      & 74.0 & 73.6 & 72.6 & 69.4 & 66.0 & 57.1 & 73.9 & 71.9 & 68.0 & 62.2 & 52.5 & 36.4 \\
& JoAPR    & 72.8 & 71.2 & 70.4 & 67.6 & 65.3 & 57.7 & 72.1 & 69.8 & 64.1 & 59.2 & 56.1 & 47.5 \\
& NLPrompt & 74.8 & 73.4 & 72.8 & 70.3 & 68.1 & 60.5 & 74.9 & 73.5 & 71.0 & 66.0 & 59.0 & 49.3 \\
& TecoPrompt
& \textbf{75.1} & \textbf{74.4} & \textbf{73.4} & \textbf{72.0} & \textbf{70.3} & \textbf{67.4}
& \textbf{75.1} & \textbf{73.9} & \textbf{71.6} & \textbf{70.1} & \textbf{67.4} & \textbf{58.1} \\
\Xhline{1pt}

\multirow{5}{*}{Caltech101}
& CoOp     & 86.4 & 81.0 & 76.7 & 70.9 & 61.3 & 46.9 & 84.9 & 75.2 & 62.9 & 49.4 & 33.6 & 20.3 \\
& GCE      & \textbf{92.0} & 90.9 & 90.8 & 89.3 & 86.7 & 79.0 & 91.3 & 91.2 & 89.7 & 85.8 & 78.2 & 62.1 \\
& JoAPR    & 90.3 & 90.5 & 89.9 & 88.3 & 86.9 & 83.9 & 90.3 & 89.3 & 88.3 & 88.7 & 85.8 & 81.9 \\
& NLPrompt & 91.7 & 91.1 & 90.8 & 89.9 & 88.3 & 86.7 & 91.6 & 91.2 & 90.2 & 89.3 & 86.2 & 81.1 \\
& TecoPrompt
& 91.5 & \textbf{91.7} & \textbf{91.1} & \textbf{90.5} & \textbf{90.1} & \textbf{89.4}
& \textbf{91.9} & \textbf{91.9} & \textbf{91.0} & \textbf{90.1} & \textbf{89.1} & \textbf{85.7} \\
\Xhline{1pt}
\end{tabular}
\end{adjustbox}
\end{table*}

\subsection{Results on Synthetic Noisy Labels}
\label{subsec:synthetic_results}

Tab.~\ref{tab:noise_results_with_Tecoprompt} summarizes performance under synthetic symmetric and asymmetric label noise across seven datasets.
The noise rate ranges from 12.5\% to 75\%, increasing in steps of 12.5\%.
Across most datasets and noise levels, TecoPrompt achieves strong performance, with particularly notable advantages emerging as the noise rate increases.
Under severe label corruption, the performance gap becomes more pronounced, demonstrating that the proposed prompt is less affected by noisy supervision.
These results indicate that TecoPrompt learns more robust prompts that maintain stable performance even under high levels of label noise.

\subsection{Results on real-world Noisy Labels}
\label{subsec:food101n}

Tab.~\ref{tab:food101n_with_Tecoprompt} reports the results on Food101N~\cite{lee2018cleannet}, a real-world dataset with naturally noisy labels. TecoPrompt achieves the highest accuracy among all compared approaches, demonstrating that the robustness observed under synthetic noise settings effectively transfers to realistic, naturally corrupted supervision.

\begin{table}[t]
\caption{Test accuracy (\%) on Food101N.}
\label{tab:food101n_with_Tecoprompt}
\centering
\fontsize{10pt}{10pt}\selectfont
\renewcommand{\arraystretch}{1.3}

\begin{tabular}{c|c|c|c|c|c}
\Xhline{1.1pt}
Method & CoOp & GCE & JoAPR & NLPrompt & TecoPrompt \\
\Xhline{0.7pt}
Accuracy & 69.50 & 71.32 & 72.57 & 76.46 & \textbf{78.67} \\
\Xhline{0.7pt}
\end{tabular}
\end{table}

\subsection{Few-shot Learning Analysis}

\begin{figure}[t]
	\centering
	\includegraphics[width=1\linewidth]{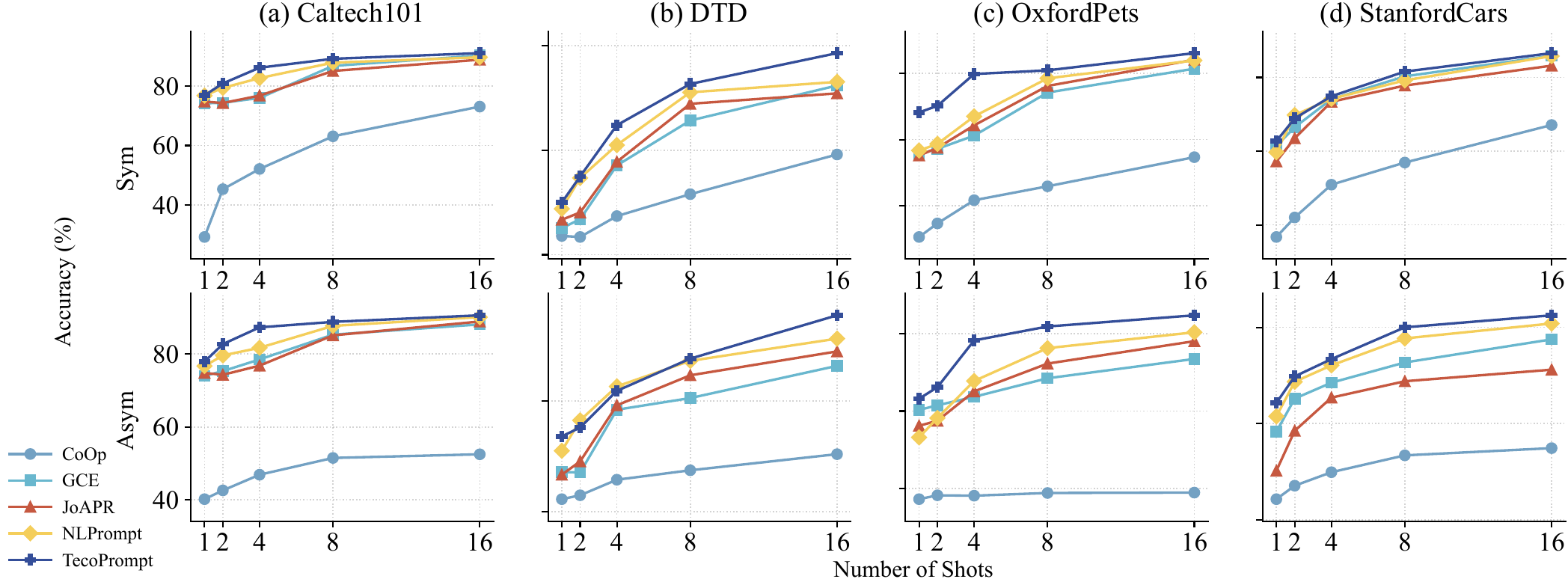}
	\caption{Few-shot robustness under label noise with different shot budgets. We vary the number of shots per class over {1, 2, 4, 8, 16}, while keeping the noise rate at 50\%.
	}
	\label{fig:figure3}
\end{figure}

To understand how data scarcity interacts with label corruption, we vary the number of shots per class in \(\{1,2,4,8,16\}\) while fixing the noise rate at 50\%. In these experiments, we set the number of training epochs to 100. The trends in Fig.~\ref{fig:figure3} show that accuracy rises steadily with more shots for all compared methods, under both symmetric and asymmetric noise. Across datasets (Caltech101, DTD, OxfordPets, StanfordCars), TecoPrompt maintains an advantage throughout, with the gap most evident at very low shot counts, indicating better robustness when both supervision quality and quantity are limited.

\subsection{Comparison of Supervision Strategies}
Tab.~\ref{tab:rewrite} reports test accuracy on Flowers102 after 100 epochs under symmetric and asymmetric label corruption at rates of 20\%, 40\%, 60\%, and 80\%. LS and ST denote Label Smoothing~\cite{szegedy2016rethinking} and Soft Targets~\cite{hinton2015distilling}, respectively. NLPrompt performs OT-based label allocation, whereas TecoPrompt further introduces EMA-smoothed confidence and a $K$-epoch stability window to enable conservative hard-label rewriting. TecoPrompt consistently achieves the best performance across all noise settings, with larger margins under severe corruption, indicating that temporally verified hard rewriting provides more reliable supervision than purely OT-based soft supervision.

\begin{table*}[t]
	\caption{Comparison of soft supervision and hard label rewriting strategies under different noise rates on Flowers102.}
	\label{tab:rewrite}
	\centering
	\renewcommand{\arraystretch}{1}
	
	\begin{tabular}{c|cccc|cccc}
		\Xhline{1.2pt}
		\multirow{2}{*}{Supervision Strategy}
		& \multicolumn{4}{c|}{Noise Rate: Sym}
		& \multicolumn{4}{c}{Noise Rate: Asym} \\
		& 20\% & 40\% & 60\% & 80\%
		& 20\% & 40\% & 60\% & 80\% \\
		\Xhline{0.6pt}
		GCE + ST       & 85.93 & 84.48 & 81.19 & 66.15 & 83.53 & 80.50 & 67.43 & 39.15 \\
		NLPrompt + ST  & 88.42 & 86.33 & 82.51 & 76.43 & 88.74 & 86.91 & 81.37 & 71.80 \\
		NLPrompt + LS  & 88.35 & 88.12 & 82.54 & 74.48 & 88.61 & 84.92 & 79.07 & 70.63 \\
		TecoPrompt     & \textbf{90.76} & \textbf{88.98} & \textbf{83.31} & \textbf{78.02}
		& \textbf{89.86} & \textbf{87.21} & \textbf{81.93} & \textbf{73.14} \\
		\Xhline{1.2pt}
	\end{tabular}
\end{table*}

\subsection{Ablation Study}

\begin{table}[t]
	\caption{Ablation studies under multiple noise ratios (\%).}
	\label{tab:ablation_with_mine}
	\centering
	\small
	\setlength{\tabcolsep}{4pt}
	\renewcommand{\arraystretch}{1.08}
	\begin{adjustbox}{max width=\textwidth,center}
		\begin{tabular}{c|c|c|c|c|c|c|c|c|c}
			\Xhline{1.2pt}
			ID & EMA & Rewrite & $K$ & Loss & 10\% & 30\% & 50\% & 70\% & Avg \\
			\Xhline{0.8pt}
			(a) &  &  & -- & -- & 87.05 & 87.03 & 85.46 & 78.22 & 84.44 \\
			(b) &  & \checkmark & -- & -- & 87.21 & 86.77 & 86.02 & 85.03 & 86.26 \\
			(c) & \checkmark &  & -- & -- & 86.05 & 85.98 & 85.34 & 81.30 & 84.67 \\
			\Xhline{0.8pt}
			(d) & \checkmark & \checkmark & -- & CE-0.3-0.5 & 87.58 & 86.91 & 85.98 & 85.36 & 86.46 \\
			(e) & \checkmark & \checkmark & -- & 0.7-0.5-0.3 & 86.70 & 85.95 & 84.35 & 58.62 & 78.91 \\
			\Xhline{0.8pt}
			(f) & \checkmark & \checkmark & 1  & -- & 82.23 & 80.85 & 78.12 & 80.15 & 80.59 \\
			(g) & \checkmark & \checkmark & 4  & -- & 86.52 & 85.75 & 85.98 & 86.02 & 86.07 \\
			(h) & \checkmark & \checkmark & 8  & -- & 87.52 & 87.19 & \textbf{86.40} & \textbf{86.72} & \textbf{86.96} \\
			(i) & \checkmark & \checkmark & 16 & -- & \textbf{87.70} & \textbf{87.61} & 85.87 & 84.74 & 86.48 \\
			\Xhline{1.2pt}
		\end{tabular}
	\end{adjustbox}
\end{table}

Tab.~\ref{tab:ablation_with_mine} summarizes ablation results of TecoPrompt on OxfordPets under symmetric label noise. All models are trained for 100 epochs under the 16-shot setting following Sec.~\ref{subsec:impl_details}. Unless otherwise specified, the default configuration uses EMA and rewriting with $K{=}8$ and loss-CE-0.5-MAE, where CE, GCE($q{=}0.5$), and MAE are applied to clean, mid, and noisy groups, respectively. Rows (a)--(c) ablate EMA and rewriting, rows (d)--(e) vary the group-wise loss design, and rows (f)--(i) study different $K$ values.

The results show that rewriting is the main driver of robustness, especially under heavy noise, while EMA alone brings smaller gains but improves stability when coupled with rewriting. The sharp collapse of (e) at 70\% noise highlights that an improper group-wise loss assignment can be detrimental in the high-noise regime, motivating the robust loss choice for the noisy group in our default design. Finally, varying $K$ indicates that moderate history yields the most stable performance: small $K$ introduces noisy corrections, whereas large $K$ makes rewriting overly conservative and reduces the frequency of corrections. The lower error ratio under larger $K$ can occasionally yield slightly better results than $K=8$, but a moderate window provides the best overall balance.

\begin{figure}[t]
	\includegraphics[width=1\linewidth]{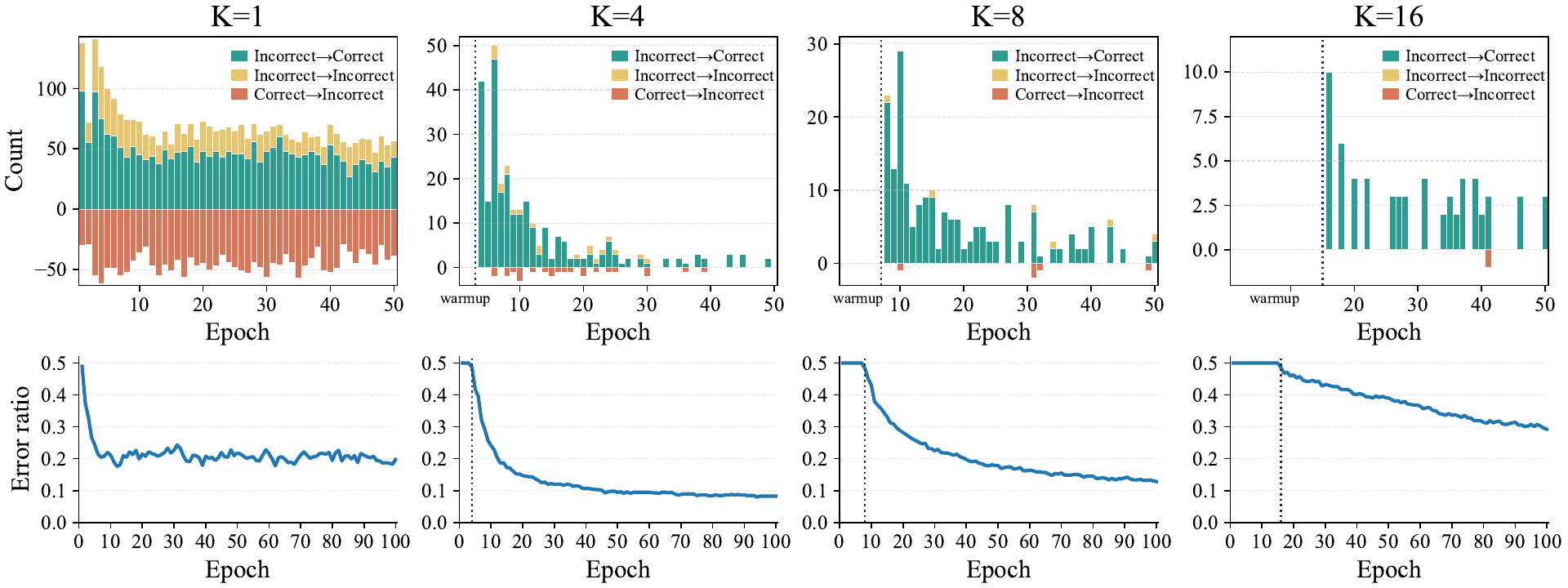}
	\caption{Effect of the temporal stability window size $K$ for conservative hard label rewriting. The top row reports the per-epoch counts of rewriting outcomes, including incorrect→correct, incorrect→incorrect, and correct→incorrect. The bottom row shows the ratio of incorrect labels in the dataset over training. The dashed line denotes the warm-up stage before rewriting is enabled.
	}
	\label{fig:figure4}
\end{figure}

Fig.~\ref{fig:figure4} provides a closer look at how $K$ affects rewriting dynamics. With a small $K$, rewriting reacts to transient fluctuations, leading to less stable corrections and a higher error ratio. Increasing $K$ suppresses such short-term label switches and reduces harmful flips, improving correction precision in a manner consistent with Theorem~\ref{thm:precision}. However, an overly large window becomes too conservative and triggers fewer effective corrections. Overall, the dynamics in Fig.~\ref{fig:figure4} explain why $K{=}8$ is a suitable default on this dataset, striking a favorable balance between correction quantity and correction precision.

\subsection{Qualitative Analysis of Label Rewriting}

On OxfordPets under 50\% label noise with 100 training epochs, TecoPrompt rewrites 234 samples, of which 222 are corrected from wrong labels to ground-truth labels, accounting for 94.87\%. Since this setting contains 296 noisy labels in total, these corrections cover 75.00\% of the noisy samples. In comparison, only 6 samples are changed from correct labels to wrong labels, and 6 from one wrong label to another wrong label, each accounting for 2.56\%. These results indicate that the temporal stability window effectively filters unreliable OT candidates and enables accurate hard rewriting.

\begin{figure}[t] 
	\centering 
	\includegraphics[width=0.9\linewidth]{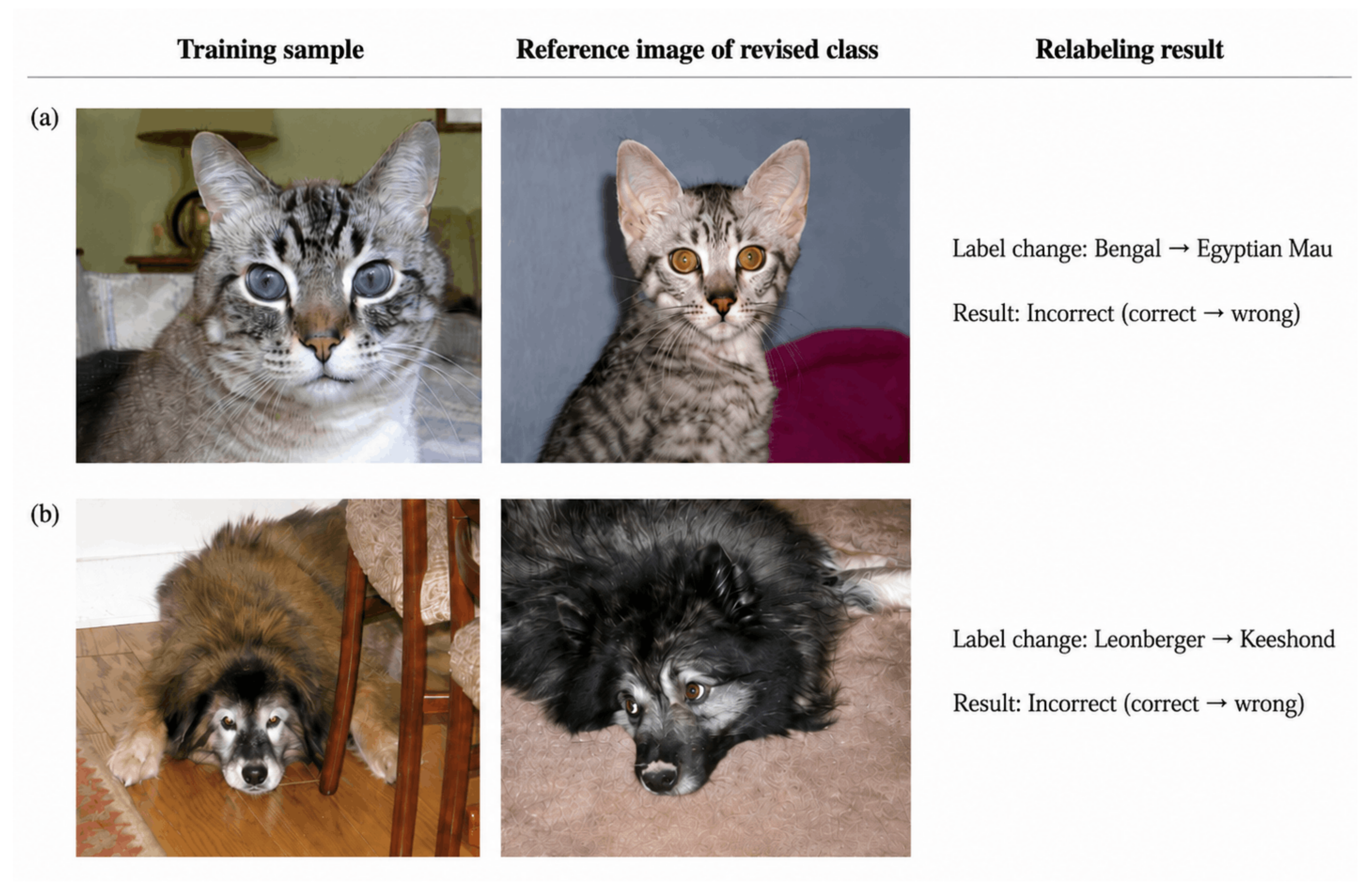} 
	\caption{
		Examples of rare incorrect label rewriting cases. The left column shows the original training samples, the middle column shows reference images from the revised classes, and the right column reports the corresponding label changes and rewriting outcomes. These cases illustrate that incorrect rewrites mainly occur between visually similar fine-grained categories.
	} 
	\label{fig:case} 
\end{figure}

Fig.~\ref{fig:case} shows the correct-to-wrong cases, where Bengal is rewritten as Egyptian Mau and Leonberger as Keeshond. These errors occur between visually similar fine-grained classes that share appearance cues such as texture, color, and facial structure. This suggests that strong inter-class similarity can occasionally lead to temporally stable but incorrect OT assignments.

\subsection{Limitations}
TecoPrompt prioritizes high-precision hard rewrites and therefore does not attempt to correct every noisy label. In practice, examples that flip frequently during training (so-called forgettable examples) typically fail the temporal verification and thus remain unmodified, which is an intentional trade-off to avoid incorrect corrections. The $K$-epoch stability window controls the precision-coverage balance and should be tuned for each dataset; empirically, $K=8$ works well across our benchmarks. Finally, the method requires maintaining historical information and EMA updates, which introduce additional overhead. However, this cost remains modest in practice; on Caltech101 under the 16-shot setting, it requires only 0.20 seconds more per epoch than NLPrompt on a single RTX 3080 Ti.

\section{Conclusion}

We propose TecoPrompt, a robust prompt-learning framework for vision--language models under noisy supervision. It stabilizes OT-based label candidates through EMA confidence gating and performs conservative rewriting only when candidates remain consistent within a $K$-epoch temporal window, reducing confirmation bias and pseudo-label drift. A group-specific objective further applies tailored losses to clean, mid, and noisy subsets. Experiments across diverse noise types and severities demonstrate consistent robustness gains, validating temporal-conservative label correction for noise-tolerant prompt tuning.

\section*{Acknowledgments.}
This work was supported in part by the National Natural Science Foundation of China (62473276, 62573309), in part by the Natural Science Foundation of Jiangsu Province (BK20241918), and in part by the Research Fund of Horizon Robotics (H230666).

\appendix
\bibliographystyle{splncs04}
\bibliography{main}

%
%

\end{document}


\appendix

\begin{center}
    {\Large \bfseries TecoPrompt: Temporal-Conservative Prompt Learning for Vision-Language Models\par}
    {\large \bfseries Supplementary Material\par}
\end{center}

\section{Additional Experiments}

\setcounter{table}{4}
\setcounter{figure}{5}

\subsection{Generalization of TecoPrompt}

Notably, our method is not specific to CoOp, and can be readily extended to other prompt learning frameworks, such as PromptSRC and MaPLe. As shown in Tab.~\ref{tab:generalization}, we further validate its effectiveness on the UCF101 dataset, where integrating our method with both PromptSRC and MaPLe consistently yields excellent performance. These results suggest that our approach is generally applicable across different prompt learning paradigms and possesses strong robustness and transferability.

\begin{table}
\caption{The generalization of TecoPrompt.}
\label{tab:generalization}
\centering
\fontsize{8pt}{10pt}\selectfont
\renewcommand{\arraystretch}{1.1}

\begin{tabular}{c|cccccc}
\Xhline{1.2pt}
Method/Noise Ratio
& 12.5\% & 25.0\% & 37.5\% & 50.0\% & 62.5\% & 75.0\% \\
\Xhline{0.6pt}
MaPLe        & 80.25 & 79.16 & 75.27 & 71.10 & 60.43 & 53.14 \\
MaPLe+Ours   & \textbf{81.96} & \textbf{81.82} & \textbf{80.65} & \textbf{79.57} & \textbf{78.68} & \textbf{76.95} \\
\Xhline{0.6pt}
PromptSRC    & 82.82 & 80.85 & 78.13 & 75.37 & 71.39 & 63.06 \\
PromptSRC+Ours & \textbf{83.62} & \textbf{82.51} & \textbf{81.90} & \textbf{80.47} & \textbf{79.89} & \textbf{77.14} \\
\Xhline{1.2pt}
\end{tabular}
\end{table}

\subsection{Experiments on SUN397}
In Tab.~\ref{tab:sun397}, TecoPrompt consistently delivers the best performance on SUN397 under both symmetric and asymmetric label noise. Our method shows clear and consistent improvements across all noise rates. These results further verify the robustness and effectiveness of our approach on challenging noisy-label settings.

\begin{table}
\caption{Test accuracy(\%) on SUN397.}
\label{tab:sun397}
\centering
\fontsize{9pt}{8pt}\selectfont
\renewcommand{\arraystretch}{1.3}

\begin{adjustbox}{width=\linewidth}

\begin{tabular}{c|c|cccccc|cccccc}
\Xhline{1.2pt}
\multirow{2}{*}{Dataset} & \multirow{2}{*}{Method}
& \multicolumn{6}{c|}{Noise Rate: Sym}
& \multicolumn{6}{c}{Noise Rate: Asym} \\

& & 12.5\% & 25.0\% & 37.5\% & 50.0\% & 62.5\% & 75.0\%
& 12.5\% & 25.0\% & 37.5\% & 50.0\% & 62.5\% & 75.0\% \\

\Xhline{0.6pt}

\multirow{4}{*}{SUN397}
& CoOp     & 65.50& 62.9 & 59.3 & 55.5 & 48.3 & 37.8 & 63.5 & 56.1 & 45.5 & 33.8 & 22.1 & 11.4 \\
& GCE      & 67.6 & 66.3 & 65.4 & 64.2 & 62.0 & 59.2 & 68.4 & 66.4 & 63.8 & 60.0 & 53.6 & 43.8 \\
& NLPrompt & 68.4 & 67.5 & 66.4 & 64.8 & 64.1 & 61.7 & 68.7 & 67.5 & 66.1 & 64.0 & 61.4 & 53.0 \\
& TecoPrompt
& \textbf{68.9} & \textbf{68.5} & \textbf{68.1} & \textbf{67.3} & \textbf{66.0} & \textbf{65.6}
& \textbf{69.4} & \textbf{68.3} & \textbf{67.4} & \textbf{66.1} & \textbf{65.1} & \textbf{60.1} \\

\Xhline{1.2pt}
\end{tabular}
\end{adjustbox}
\end{table}

\subsection{Additional Ablation Study}
Table~\ref{tab:table5} further evaluates the effects of EMA, the temporal window size $K$, and the loss design on EuroSAT. The results show that EMA or temporal consistency alone provides limited gains, while their combination significantly improves robustness, with $\beta=0.8$ and $K=8$ achieving the best average accuracy. This confirms that EMA smoothing stabilizes confidence estimation and the $K$-epoch window helps avoid unreliable rewriting. Meanwhile, the performance drop with smaller or larger $K$ indicates that a moderate window better balances correction precision and coverage. In addition, the proposed tri-group loss outperforms CE, GCE, and MAE, demonstrating the benefit of applying different losses to samples with different reliability levels.
\begin{table}[t]
\caption{Additional Ablation study on the EuroSAT dataset.}
\label{tab:table5}
\centering
\renewcommand{\arraystretch}{0.9}
\begin{tabular}{c|c|c|ccccc}
\Xhline{0.8pt}
\multirow{2}{*}{$\beta$(EMA)} 
& \multirow{2}{*}{$K$} 
& \multirow{2}{*}{Loss}
& \multicolumn{4}{c}{Noise Rate}
& \multirow{2}{*}{Average} \\
& & & 10\% & 30\% & 50\% & 70\% & \\
\Xhline{0.6pt}

\textemdash & $\infty$ & Ours & 78.23 & 75.84 & 41.15 & 19.29 & 53.63 \\
0.8 & $\infty$ & Ours & 79.34 & 75.92 & 51.76 & 11.88 & 54.73 \\
\textemdash & 8 & Ours & 78.69 & 76.15 & 65.61 & 20.36 & 60.20 \\
\Xhline{0.6pt}

\multirow{3}{*}{0.4}
& 4  & Ours & 77.90 & 75.95 & 65.59 & 40.33 & 64.94 \\
& 8  & Ours & 80.25 & 76.87 & 68.58 & 42.68 & 67.09 \\
& 12 & Ours & 79.32 & 79.46 & 67.11 & 33.15 & 64.76 \\
\Xhline{0.6pt}

\multirow{3}{*}{0.6}
& 4  & Ours & 77.73 & 75.45 & 66.41 & 27.83 & 61.86 \\
& 8  & Ours & 80.24 & 76.42 & 68.62 & 31.57 & 64.21 \\
& 12 & Ours & 80.75 & 77.04 & 68.01 & 25.05 & 62.71 \\
\Xhline{0.6pt}

\multirow{3}{*}{0.8}
& 4  & Ours & 79.13 & 75.12 & 67.24 & 42.85 & 66.08 \\
& 8  & Ours & 80.23 & 77.03 & \textbf{70.53} & \textbf{43.83} & \textbf{67.91} \\
& 12 & Ours & \textbf{80.94} & \textbf{77.51} & 68.65 & 32.45 & 64.89 \\

\Xhline{0.6pt}

\multirow{3}{*}{0.8}
& 8 & CE & 79.28 & 72.33 & 58.76 & 33.32 & 60.92 \\
& 8 & GCE & 78.89 & 74.47 & 65.67 & 35.61 & 63.66 \\
& 8 & MAE & 76.33 & 72.03 & 45.37 & 21.46 & 53.80 \\

\Xhline{1.2pt}
\end{tabular}
\end{table}

\section{Theoretical Analysis of Temporal-Consistency-Triggered OT Label Rewriting}
\label{sec:theory_appendix}

In this section, we analyze when a temporally stable OT assignment can serve as a reliable rewriting candidate in TecoPrompt, and how enlarging the consistency window $K$ improves rewrite precision.

\subsection{Notation and Definitions}

Let $\mathcal{D}=\{(x_i,\tilde y_i)\}_{i=1}^N$ be a dataset with observed labels $\tilde y_i\in\{1,\dots,C\}$. Each sample has an unknown ground-truth label $y_i^\star$. All probabilities are taken with respect to a uniformly random index $i$ together with any algorithmic randomness.

Define the noise indicator and the noise rate as
\begin{equation}
	Z_i \triangleq \mathbf{1}[\tilde y_i \neq y_i^\star],
	\qquad
	\eta \triangleq \mathbb{P}(Z_i=1).
	\label{eq:noise_def}
\end{equation}

At epoch $s$, let $\hat y_i^{(s)}\in\{1,\dots,C\}$ denote the OT-induced pseudo-label of sample $i$, and let $c_i^{(s)}\in[0,1]$ be the corresponding confidence score.

Define the flip indicator of the OT pseudo-label trajectory by
\begin{equation}
	d_i^{(s)} \triangleq \mathbf{1}\!\left[\hat y_i^{(s+1)} \neq \hat y_i^{(s)}\right].
	\label{eq:flip_indicator}
\end{equation}

Fix a window $[t_0,\,t_0+T_w-1]$ and define the flip rate as
\begin{equation}
	\phi_i \triangleq \frac{1}{T_w}\sum_{s=t_0}^{t_0+T_w-1} d_i^{(s)} \in [0,1].
	\label{eq:flip_rate_def}
\end{equation}

Given a threshold $\phi_0\in(0,1)$, define the OT-forgettability indicator by
\begin{equation}
	F_i \triangleq \mathbf{1}[\phi_i > \phi_0],
	\qquad
	\tau \triangleq \mathbb{P}(F_i=1).
	\label{eq:flip_def}
\end{equation}

We refer to samples with $F_i=0$ as \emph{OT-unforgettable} and to samples with $F_i=1$ as \emph{OT-forgettable}. Unlike the classical notion of forgetting, which is defined in terms of correctness with respect to $y_i^\star$, both $\phi_i$ and $F_i$ are directly observable from the OT pseudo-label trajectory.

The pair $(Z_i,F_i)$ induces the partition
\begin{equation}
	\begin{array}{c|c@{\hspace{12pt}}c}
		& F_i=0,\ \text{OT-unforgettable} & F_i=1,\ \text{OT-forgettable}\\
		\hline
		Z_i=0,\ \text{clean} & \mathcal{S}_{c,u} & \mathcal{S}_{c,f}\\
		Z_i=1,\ \text{noisy} & \mathcal{S}_{n,u} & \mathcal{S}_{n,f}
	\end{array}
	\label{eq:2x2}
\end{equation}

To characterize the coupling between noise and OT-forgettability, define
\begin{equation}
	\rho_u \triangleq \mathbb{P}(F_i=0\mid Z_i=1)\in[0,1]
	\qquad
	(\eta>0).
	\label{eq:rho_u_def}
\end{equation}

Then the block masses are given by
\begin{equation}
	\pi_{n,u}=\eta\rho_u,
	\quad
	\pi_{n,f}=\eta(1-\rho_u),
	\quad
	\pi_{c,f}=\tau-\eta(1-\rho_u),
	\quad
	\pi_{c,u}=1-\tau-\eta\rho_u.
	\label{eq:block_masses}
\end{equation}

Define the $K$-epoch stable-mismatch event by
\begin{equation}
	\mathcal{E}_K(i) \triangleq \{\hat y_i^{(t-K+1)}=\cdots=\hat y_i^{(t)} \neq \tilde y_i\}.
	\label{eq:EK_def}
\end{equation}

On $\mathcal{E}_K(i)$, define the stable label as $\bar y_i \triangleq \hat y_i^{(t)}$.

Define the $K$-epoch confidence gate by
\begin{equation}
	\mathcal{D}_K(i) \triangleq \{c_i^{(t-K+1)}\ge\theta,\ \dots,\ c_i^{(t)}\ge\theta\}.
	\label{eq:DK_def}
\end{equation}

Finally, define the rewrite-candidate event by
\begin{equation}
	\mathcal{A}_K(i) \triangleq \mathcal{E}_K(i) \cap \mathcal{D}_K(i).
	\label{eq:AK_def}
\end{equation}

\subsection{Assumptions}

\begin{assumption}
	\label{assump:gap}
	A one-step stability separation holds: conditional on past stability, OT-unforgettable samples retain the same OT pseudo-label with probability at least $p_u$, whereas OT-forgettable samples do so with probability at most $p_f$. Specifically, there exist constants $0<p_f<p_u<1$ such that for every $s\in\{t-K+1,\dots,t-1\}$,
	\begin{equation}
		\mathbb{P}\!\left(\hat y^{(s+1)}=\hat y^{(s)} \,\middle|\, F=0,\ \hat y^{(t-K+1)}=\cdots=\hat y^{(s)}\right) \ge p_u,
		\label{eq:gap_unf}
	\end{equation}
	whereas
	\begin{equation}
		\mathbb{P}\!\left(\hat y^{(s+1)}=\hat y^{(s)} \,\middle|\, F=1,\ \hat y^{(t-K+1)}=\cdots=\hat y^{(s)}\right) \le p_f.
		\label{eq:gap_for}
	\end{equation}
	where $p_u$ and $p_f$ denote the corresponding one-step stability bounds.
\end{assumption}

\begin{assumption}
	\label{assump:conf_gate}
	The confidence gate preserves stable candidates on the two OT-unforgettable blocks with probabilities bounded below by $q_{c,u}$ and $q_{n,u}$. Concretely, there exist constants $q_{c,u},q_{n,u}\in(0,1]$ such that
	\begin{equation}
		\mathbb{P}(\mathcal{D}_K \mid \mathcal{S}_{c,u},\ \mathcal{E}_K) \ge q_{c,u},
		\qquad
		\mathbb{P}(\mathcal{D}_K \mid \mathcal{S}_{n,u},\ \mathcal{E}_K) \ge q_{n,u}.
		\label{eq:conf_gate}
	\end{equation}
	This assumption captures the idea that stable candidates are not frequently discarded by confidence thresholding.
\end{assumption}

\begin{assumption}
	\label{assump:correct}
	On the target block $\mathcal{S}_{n,u}$, rewritten labels are correct up to a residual error $\varepsilon_\star$. That is, there exists $\varepsilon_\star\in[0,1)$ such that
	\begin{equation}
		\mathbb{P}(\bar y = y^\star \mid \mathcal{S}_{n,u},\ \mathcal{A}_K) \ge 1-\varepsilon_\star.
		\label{eq:correct}
	\end{equation}
	where $\varepsilon_\star$ denotes the residual error rate on $\mathcal{S}_{n,u}$.
\end{assumption}

\begin{assumption}
	\label{assump:stable_mismatch}
	The stable-mismatch event is exponentially most likely on $\mathcal{S}_{n,u}$. Specifically, there exist constants
	\begin{equation}
		0<p_f<p_{c,u}<p_{n,u}<1,
		\qquad
		\varepsilon_c,\varepsilon_n\in[0,1),
		\label{eq:sep_constants}
	\end{equation}
	such that for every $K\ge2$,
	\begin{equation}
		\mathbb{P}(\mathcal{E}_K \mid \mathcal{S}_{n,u}) \ge (1-\varepsilon_n)p_{n,u}^{K-1},
		\qquad
		\mathbb{P}(\mathcal{E}_K \mid \mathcal{S}_{c,u}) \le \varepsilon_c p_{c,u}^{K-1},
		\label{eq:stable_mismatch_unf}
	\end{equation}
	and
	\begin{equation}
		\mathbb{P}(\mathcal{E}_K \mid \mathcal{S}_{c,f}) \le p_f^{K-1},
		\qquad
		\mathbb{P}(\mathcal{E}_K \mid \mathcal{S}_{n,f}) \le p_f^{K-1}.
		\label{eq:stable_mismatch_for}
	\end{equation}
	Thus, $p_{n,u}$ is the dominant rate.
\end{assumption}

\subsection{Main Results and Proofs}

\begin{lemma}
	\label{lem:cons_likelihood}
	The following lemma quantifies the stability gap between OT-unforgettable and OT-forgettable samples over a $K$-epoch window.
	
	Under Assumption~\ref{assump:gap}, for any $K\ge2$,
	\begin{equation}
		\begin{aligned}
			\mathbb{P}(\hat y^{(t-K+1)}=\cdots=\hat y^{(t)} \mid F=0)
			&\ge p_u^{K-1}, \\
			\mathbb{P}(\hat y^{(t-K+1)}=\cdots=\hat y^{(t)} \mid F=1)
			&\le p_f^{K-1}.
		\end{aligned}
		\label{eq:cons_bounds}
	\end{equation}
\end{lemma}

\begin{proof}
	For $s\in\{t-K+1,\dots,t-1\}$, let $G_s\triangleq\{\hat y^{(s+1)}=\hat y^{(s)}\}$. Then
	\[
	\{\hat y^{(t-K+1)}=\cdots=\hat y^{(t)}\}=\bigcap_{s=t-K+1}^{t-1} G_s.
	\]
	If $F=0$, the chain rule gives
	{\small
		\begin{equation}
			\mathbb{P}\!\left(\bigcap_{s=t-K+1}^{t-1} G_s \middle| F=0\right)
			=
			\prod_{s=t-K+1}^{t-1}
			\mathbb{P}\!\left(G_s \middle| F=0,\ G_{t-K+1},\dots,G_{s-1}\right).
			\label{eq:chain_rule_unf_revised}
		\end{equation}
	}
	Since $G_{t-K+1},\dots,G_{s-1}$ imply $\hat y^{(t-K+1)}=\cdots=\hat y^{(s)}$, each factor is at least $p_u$ by \eqref{eq:gap_unf}. As there are $K-1$ such factors, we obtain
	\[
	\mathbb{P}(\hat y^{(t-K+1)}=\cdots=\hat y^{(t)} \mid F=0)\ge p_u^{K-1}.
	\]
	The case $F=1$ is analogous: by \eqref{eq:gap_for}, each factor is at most $p_f$, so
	\[
	\mathbb{P}(\hat y^{(t-K+1)}=\cdots=\hat y^{(t)} \mid F=1)\le p_f^{K-1}.
	\]
\end{proof}

\begin{theorem}
	\label{thm:rewrite_correctness}
	The following result reduces rewrite precision to the proportion of $\mathcal{S}_{n,u}$ within the candidate set $\mathcal{A}_K$.
	
	Under Assumption~\ref{assump:correct},
	\begin{equation}
		\mathbb{P}(\bar y=y^\star \mid \mathcal{A}_K)
		\ge
		(1-\varepsilon_\star)\,\mathbb{P}(\mathcal{S}_{n,u}\mid \mathcal{A}_K).
		\label{eq:rewrite_lb}
	\end{equation}
\end{theorem}

\begin{proof}
	Since $\mathcal{S}_{c,u}$, $\mathcal{S}_{c,f}$, $\mathcal{S}_{n,u}$, and $\mathcal{S}_{n,f}$ form a partition, the law of total probability under $\mathcal{A}_K$ gives
	\begin{equation}
		\mathbb{P}(\bar y=y^\star \mid \mathcal{A}_K)
		=
		\sum_{B\in\{\mathcal{S}_{c,u},\mathcal{S}_{c,f},\mathcal{S}_{n,u},\mathcal{S}_{n,f}\}}
		\mathbb{P}(\bar y=y^\star \mid \mathcal{A}_K,B)\,\mathbb{P}(B\mid \mathcal{A}_K).
		\label{eq:rewrite_total_prob_revised}
	\end{equation}
	Retaining only the nonnegative term corresponding to $\mathcal{S}_{n,u}$ yields
	\[
	\mathbb{P}(\bar y=y^\star \mid \mathcal{A}_K)
	\ge
	\mathbb{P}(\bar y=y^\star \mid \mathcal{A}_K,\mathcal{S}_{n,u})\,
	\mathbb{P}(\mathcal{S}_{n,u}\mid \mathcal{A}_K).
	\]
	Assumption~\ref{assump:correct} bounds the first factor from below by $1-\varepsilon_\star$, proving \eqref{eq:rewrite_lb}.
\end{proof}

\begin{theorem}
	\label{thm:nu_under_AK}
	The following theorem shows that, as $K$ increases, the candidate set $\mathcal{A}_K$ becomes increasingly concentrated on the noisy OT-unforgettable block.
	
	Under Assumptions~\ref{assump:stable_mismatch} and \ref{assump:conf_gate}, for every $K\ge2$,
	{\small
		\begin{equation}
			\mathbb{P}(\mathcal{S}_{n,u}\mid \mathcal{A}_K)
			\ge
			\frac{q_{n,u}(1-\varepsilon_n)\pi_{n,u}}{q_{n,u}(1-\varepsilon_n)\pi_{n,u}+\varepsilon_c\left(\frac{p_{c,u}}{p_{n,u}}\right)^{K-1}\pi_{c,u}+\left(\frac{p_f}{p_{n,u}}\right)^{K-1}(\pi_{c,f}+\pi_{n,f})}.
			\label{eq:nu_lowerbound}
		\end{equation}
	}
	Moreover, the right-hand side is nondecreasing in $K$ and converges to $1$ as $K\to\infty$.
\end{theorem}

\begin{proof}
	By Bayes' rule and the four-block partition,
	{\scriptsize
		\begin{equation}
			\mathbb{P}(\mathcal{S}_{n,u}\mid \mathcal{A}_K)
			=
			\frac{\mathbb{P}(\mathcal{A}_K\mid \mathcal{S}_{n,u})\pi_{n,u}}{\mathbb{P}(\mathcal{A}_K\mid \mathcal{S}_{n,u})\pi_{n,u}+\mathbb{P}(\mathcal{A}_K\mid \mathcal{S}_{c,u})\pi_{c,u}+\mathbb{P}(\mathcal{A}_K\mid \mathcal{S}_{c,f})\pi_{c,f}+\mathbb{P}(\mathcal{A}_K\mid \mathcal{S}_{n,f})\pi_{n,f}}.
			\label{eq:partition_step_revised}
		\end{equation}
	}
	For the target block,
	\[
	\mathbb{P}(\mathcal{A}_K\mid \mathcal{S}_{n,u})
	=
	\mathbb{P}(\mathcal{D}_K\mid \mathcal{S}_{n,u},\mathcal{E}_K)\,
	\mathbb{P}(\mathcal{E}_K\mid \mathcal{S}_{n,u})
	\ge
	q_{n,u}(1-\varepsilon_n)p_{n,u}^{K-1},
	\]
	where we used Assumptions~\ref{assump:conf_gate} and \ref{assump:stable_mismatch}. For the remaining blocks, the inclusion $\mathcal{A}_K\subseteq\mathcal{E}_K$ gives
	\[
	\mathbb{P}(\mathcal{A}_K\mid \mathcal{S}_{c,u})\le \varepsilon_c p_{c,u}^{K-1},
	\qquad
	\mathbb{P}(\mathcal{A}_K\mid \mathcal{S}_{c,f})\le p_f^{K-1},
	\qquad
	\mathbb{P}(\mathcal{A}_K\mid \mathcal{S}_{n,f})\le p_f^{K-1}.
	\]
	Substituting these bounds into \eqref{eq:partition_step_revised} yields
	{\small
		\begin{equation}
			\mathbb{P}(\mathcal{S}_{n,u}\mid \mathcal{A}_K)
			\ge
			\frac{q_{n,u}(1-\varepsilon_n)p_{n,u}^{K-1}\pi_{n,u}}{q_{n,u}(1-\varepsilon_n)p_{n,u}^{K-1}\pi_{n,u}+\varepsilon_c p_{c,u}^{K-1}\pi_{c,u}+p_f^{K-1}(\pi_{c,f}+\pi_{n,f})}.
			\label{eq:nu_before_divide_revised}
		\end{equation}
	}
	Dividing the numerator and denominator by $p_{n,u}^{K-1}$ gives \eqref{eq:nu_lowerbound}. Since
	\[
	0<\frac{p_{c,u}}{p_{n,u}}<1,
	\qquad
	0<\frac{p_f}{p_{n,u}}<1,
	\]
	the two contamination terms decrease monotonically to zero as $K$ increases, whereas the leading term $q_{n,u}(1-\varepsilon_n)\pi_{n,u}$ remains constant. Hence, the lower bound is nondecreasing in $K$ and converges to $1$.
\end{proof}

\begin{theorem}
	\label{thm:precision_app}
	Combining the previous two results yields an explicit lower bound on rewrite precision.
	
	Under Theorem~\ref{thm:rewrite_correctness} and Assumptions~\ref{assump:conf_gate}--\ref{assump:stable_mismatch},
	{\scriptsize
		\begin{equation}
			\mathbb{P}(\bar y = y^\star \mid \mathcal{A}_K)
			\ge
			(1-\varepsilon_\star)
			\frac{q_{n,u}(1-\varepsilon_n)\pi_{n,u}}{q_{n,u}(1-\varepsilon_n)\pi_{n,u}+\varepsilon_c\left(\frac{p_{c,u}}{p_{n,u}}\right)^{K-1}\pi_{c,u}+\left(\frac{p_f}{p_{n,u}}\right)^{K-1}(\pi_{c,f}+\pi_{n,f})}.
			\label{eq:precision_final}
		\end{equation}
	}
	Hence, the lower bound on rewrite precision is nondecreasing in $K$ and converges to $1-\varepsilon_\star$ as $K\to\infty$.
\end{theorem}

\begin{proof}
	Theorem~\ref{thm:rewrite_correctness} gives
	\[
	\mathbb{P}(\bar y = y^\star \mid \mathcal{A}_K)
	\ge
	(1-\varepsilon_\star)\,\mathbb{P}(\mathcal{S}_{n,u}\mid \mathcal{A}_K),
	\]
	and Theorem~\ref{thm:nu_under_AK} lower-bounds the second factor by the fraction in \eqref{eq:nu_lowerbound}. Substituting this bound proves \eqref{eq:precision_final}. The monotonicity and limiting value follow immediately from Theorem~\ref{thm:nu_under_AK}, since the prefactor $(1-\varepsilon_\star)$ is independent of $K$.
\end{proof}

%
%